# Using Wikipedia to Boost SVD Recommender Systems

Gilad Katz[1,2]
katzgila@bgu.ac.il

Guy Shani[1]
shanigu@bgu.ac.il

Bracha Shapira[1,2]
bshapira@bgu.ac.il

Lior Rokach[1,2]
liorrk@bgu.ac.il

[1] Department of Information Systems Engineering, Ben Gurion University, Israel
[2] Deutsche Telekom laboratories at Ben Gurion University, Israel

## ABSTRACT
Singular Value Decomposition (SVD) has been used successfully in recent years in the area of recommender systems. In this paper we present how this model can be extended to consider both user ratings and information from Wikipedia. By mapping items to Wikipedia pages and quantifying their similarity, we are able to use this information in order to improve recommendation accuracy, especially when the sparsity is high. Another advantage of the proposed approach is the fact that it can be easily integrated into any other SVD implementation, regardless of additional parameters that may have been added to it. Preliminary experimental results on the MovieLens dataset are encouraging.

## 1. INTRODUCTION
An important task of recommender systems is the prediction of unknown user-item ratings. In recent years, Singular Value Decomposition (SVD) [1] has become a popular tool for that task. The effectiveness if the SVD method has been demonstrated during the Netflix Challenge, where it was a crucial component of the winning algorithm.

The field of item ratings predictions faces two major difficulties – data sparsity and the cold start problem. The user-rating matrix is in many cases spars, i.e., contains very little known ratings. It is therefore difficult to train algorithms based solely on these ratings. The cold start problem is similar, but focuses on new items, where ratings are difficult to predict before sufficient data has been collected [2, 3].

We have previously [4] shown how data from Wikipedia can be used to compute item-item similarities. Using theses similarities we can propagate user ratings over a small set of items to other, similar items, thus enriching the user profile. We have shown that the enriched user profile can be used by classic CF algorithms to compute user-user correlations and improve rating predictions [4].

In this paper we leverage the Wikipedia based item similarities in the context of SVD algorithms. We show various methods for integrating this data, starting from adjusting the learning rate, through a mixture of two different SVD models, and ending with an integration of item similarities directly into the latent features.

We evaluate the performance of the proposed methods over the MovieLens dataset, simulating various sparsity levels. We show that many of the proposed methods outperform the regular SVD algorithm, particularly when data sparsity is high.

## 2. Background
### 2.1 Integration of external information in RS
In collaborative filtering (CF), rating predictions are computed using existing user-item rating set, leveraging the similarity in ratings patterns of similar users. It is now widely agreed that, given sufficient density of the user-item rating matrix, CF methods achieve the best prediction accuracy, without requiring any external data sources. In many cases however, such as when the system is relatively new or when the user population is relatively small or when users tend to rate only a handful of items, CF methods suffer from sparsity and the cold-start problem and might be augmented using such external data.

Indeed, the idea of integrating external sources to boost recommender systems was already explored in previous studies, differing on the data sources they use and on the method used to integrate these sources. For example, [5] uses organizational social networks in order to compute similarities among users. Possible indicators of similarities include attending the same events, co-authored paper, and being members of the same projects. The similarities are used to create profiles which are then integrated into a web based recommender system.

One can also use an item's content information from external sources. This content may include the genres, actors or director of a movie item, in order to calculate item correlations. For example, [6] integrates information from a set of knowledge sources, including Wikipedia, and generates a graph of linguistic terms. The generated model enables identifying similar items for any given item. However (as opposed to our work), [6] uses only selected terms and does not use all the user-generated text. In addition, they ignore Wikipedia attributes, such as categories. Other methods that use Wikipedia include [7], where semantic relations extraction was used in order to represent domain resources as a set of interconnected Wikipedia articles. For this purpose, the authors used only the links of the Wikipedia pages and ignored other features such as text and page categories.

The use of Wikipedia features to boost recommender systems was also previously explored in [4], where we show how it can be used to boost item-item based collaborative filtering. This was done by generating "artificial ratings" for additional items, based on a weighted average of their Wikipedia similarity to other items rated by the user. These ratings were then used to reduce the sparsity of the data.

### 2.2 SVD
SVD is a well-known matrix factorization method that gained much popularity following its successful application in the Netflix

Challenge [8]. It utilizes latent relations between users and items and represents them in a latent factor space of dimensionality.

SVD computes predictions for unknown items using the following base model [9]:

$$\hat{r}_{u,i} = \mu + b_i + b_u + q_i^T \cdot p_u$$

Where $\mu$ is the average rating, $b_u$ and $b_i$ are the item and user biases (respectively), and $q_i$ and $p_u$ are vectors of latent features for the item and the user (respectively). The parameters of this model can be trained by a stochastic gradient descent algorithm:

$$b_i = b_i + \gamma \cdot e \cdot \lambda \cdot b_i$$
$$b_u = b_u + \gamma \cdot e \cdot \lambda \cdot b_u$$
$$q_i = q_i + \gamma(e_{ui} \cdot p_u - \lambda \cdot q_i)$$
$$p_u = p_u + \gamma(e_{ui} \cdot q_i - \lambda \cdot p_u)$$

Where $\gamma$ is the learning step size and $\lambda$ is a regularization parameter.

In recent years many versions of the algorithm have been proposed. The modifications to the basic algorithm include additional parameters [10], incremental learning [11] and integration with additional recommendation methods [12].

## 3. Wikipedia-based Item Similarity

We compute item-item similarities using information on Wikipedia. We first match items to their corresponding Wikipedia pages, and then use various Wikipedia attributes to compute item-item similarities. For some of our methods, we also use the item-item similarities to generate artificial user-item ratings.

### 3.1 Assigning Wikipedia Pages to Items

Successfully matching items to their corresponding Wikipedia pages is crucial to the proposed method. The matching was done by generating several variations of the name for each item, looking for exact matches in Wikipedia and choosing the page with most categories containing certain keywords. For additional details see [4]. Using this method we were successful in matching 90% of items in the MovieLens dataset that contains 943 users and 1682 items.

### 3.2 Calculating the Similarity for Each Pair of Items

In [4] we define and evaluate three types of item-item similarities: text, categories and links. The text similarity is calculated using the cosine measure of the TF-IDF vectors [13] of the text of the Wikipedia pages, while the categories and links similarities are calculated by counting the number of categories or links each pair of items had in common. Our evaluation demonstrated that for the MovieLens dataset, the use of the category similarity produced results that were close to the optimal combination of similarities. For this reason we focus on category similarity only in this paper, which will be dubbed item similarity.

Over 2500 distinct categories were identified for the dataset, ranging from highly used categories such as "Films of the 1990s" to rarer one such as "Cloning in Fiction" (one possible direction for further investigation is assigning weights to categories based on their commonality).

### 3.3 Generating Artificial Ratings

The artificial ratings, which were used in some of the methods presented below, were generated as follows; for each missing user rating, we use all the items that were rated by the user and have a Wikipedia similarity to the item. The artificial rating is the calculated using the following formula: $\tilde{r}_{ui} = \frac{\sum_{j \in K} R_{uj} \cdot sim(i,j)}{\sum_{j \in K} sim(i,j)}$

Clearly, this method does not assign ratings to all unknown items because many items were not similar to any item rated by the user $u$. Furthermore, some items could not be successfully assigned to Wikipedia pages. Note that these artificial ratings are not presented to users as predictions, but are only used as an intermediate step in our algorithms.

## 4. Adapting SVD to Wikipedia-Based Item Similarities

We now present several possible modifications to the SVD model that allow us to leverage data from Wikipedia in computing user-item rating predictions. These modified models will be later evaluated and compared.

There are two possible approaches for integrating knowledge from Wikipedia in the model. The first is to use the artificial ratings directly (models a-c), and the second is to use the item-item similarity instead (models d-f).

### a) Smaller learning step size for artificial ratings

We first use both the true user-item ratings and the artificial ratings to train an SVD model. The number of artificial ratings is typically much larger than the number of true ratings, and therefore creates a bias towards predicting the artificial ratings. We mitigate this effect by using a smaller step size $\tilde{\gamma}$ when updating the model parameters using an error over the artificial ratings.

In order to set the value of $\tilde{\gamma}$ we count the number of original and artificial ratings in the dataset. We found that the ratio of artificial to original ratings is roughly 100 to 1 and therefore set the weights appropriately: $\gamma = 0.005$ and $\tilde{\gamma} = 0.00005$.

### b) Additional parameters for artificial ratings

We now separate between learning for artificial and true ratings, by adding a new set of model parameters that capture the prediction using the artificial ratings:

$$\hat{r}_{ui} = (\mu + b_i + b_u + q_i^T p_u) + (\tilde{b}_i + \tilde{b}_u + \tilde{q}_i^T \tilde{p}_u)$$

We train the model again on all ratings, both the true and the artificial ones, but update in each case only one set of parameters; When the predictions is done for a true rating, only the true model parameters are updated, and when a predictions is done for an artificial rating, only the model parameters associated with the artificial ratings are updated.

Again, we used two different value of $\gamma$, as presented in the previous section, to compensate for the larger amount of artificial ratings.

### c) Mixture model for true and artificial ratings

In this setting we trained two separate SVD models, one for the original ratings and another for the artificial ones. During the prediction phase, a weighted average of the two predictions is used to produce the final prediction.

We've experimented with several sets of weights for combining the mixture models and found little sensitivity to this parameter. Therefore, the experiments below apply equal weight to both models.

**d) Item assistance factor**

In this setting we added a set of scalars $y_i$ to the model, where $y_i$ denotes the extent to which the other items should influence predictions for item $i$. We expect higher values of $y_i$ as we have less data for training $i$, meaning that we require more "assistance" from Wikipedia similarities for that item.

The new model is as follows:

$$\hat{r}_{u,i} = \mu + b_i + b_u + q_i^T \cdot p_u + y_i \sum \frac{sim_{ij} r_{uj}}{sim_{ij}}$$

The same $\gamma$ and $\lambda$ parameters were used for all elements of the equation.

**e) User-item assistance factor**

Clearly, an item may require more "assistance" from Wikipedia similarities in the context of certain users. Our next variation captures this:

$$\hat{r}_{u,i} = \mu + b_i + b_u + q_i^T \cdot p_u + y_{ij} \sum \frac{sim_{ij} r_{uj}}{sim_{ij}}$$

**f) Adding latent factors**

We now add additional latent factors, weighted by the item-item similarities and the true ratings of other items, avoiding the intermediate artificial rating computation. Our approach is motivated by the addition of implicit rating information in [10].

The proposed model is:

$$\hat{r}_{u,i} = \mu + b_i + b_u + q_i^T \cdot p_u^T (q_i + \frac{\sum_{j \in R(u)} sim_{ij} r_{uj}}{sim_{ij}})$$

where we add to the latent features of item $i$ a set of latent features of other items $j$ that have been rated by the user and are similar to $i$. The additional latent features are weighted by the normalized item-item similarities, and the true rating of j. This allows the model to learn which items are more influential on other items.

The model parameters are updated using:

$$p_i \leftarrow p_i + \gamma \cdot \left( e_{ui} \cdot \left( q_i + \frac{\sum_{j \in R(u)} sim_{ij} r_{uj}}{\sum_{j \in R(u)} sim_{ij}} y_j \right) - \lambda \cdot p_i \right)$$

$$y_j \leftarrow y_j + \gamma \cdot \left( e_{ui} \cdot \frac{\sum_{j \in R(u)} sim_{ij} r_{uj}}{\sum_{j \in R(u)} sim_{ij}} y_j \right) - \lambda \cdot p_i)$$

The other parameters are updated in the same manner as in the previous models.

## 5. Evaluation

We evaluate the proposed method using the MovieLens dataset. This dataset contains 943 users and 1682 items. As a baseline, we used the basic SVD model [9] described in section 2.2.

We used learning set sizes ranging between 5% and 80% of the total provided ratings, providing sparsity levels between 99.68% to 95%, respectively. The learning sets were generated by randomly choosing, for each user, a portion of her provided ratings.

As mentioned in section 4, the proposed models can be divided into two groups – those that utilize the artificial ratings and those that rely on the item-item similarity obtained from Wikipedia. The results of the first group models are presented in Figure 1 and those of the second group are presented in Figure 2. It is clear that all the models of the first group outperform the baseline classifier, while those in the second group fare much worse. All results were found to be significant using paired t-test with a confidence level of 95%. In fact, the only model to present any improvement in this group is the model utilizing the user-item similarity vector. The performance results of each model compared to the baseline is presented in Tables 1 and 2.

## 6. Conclusions

In this paper we examine the benefits of Wikipedia as an external source of information for SVD–based collaborative filtering algorithms. We examine several methods for the integration of the additional information in the SVD model, particularly the use of artificial ratings and the calculation of item-item similarity.

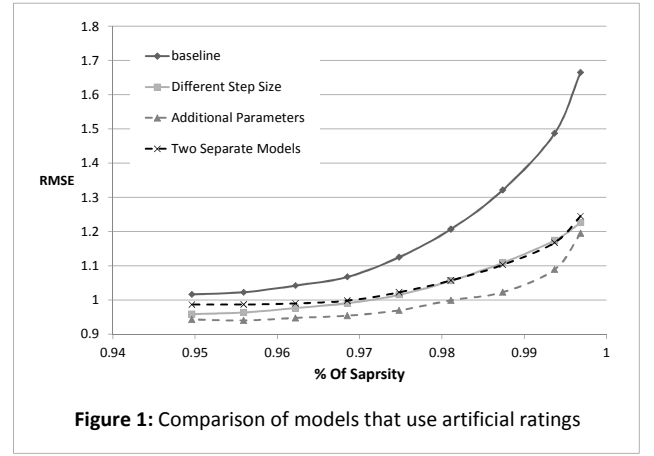

**Figure 1:** Comparison of models that use artificial ratings

Our experiments have clearly shown that the use of the artificial ratings presented in [4] produces superior results to those obtained by only calculating the similarity of items using Wikipedia, especially when sparsity is high (although the improvement is significant in all evaluated levels of sparsity).

We believe that the superiority of the models that leverage the artificial ratings is rooted in the fact that the stochastic gradient descent algorithm cannot learn accurately the additional parameters, given the very small amount of available data in these sparsity levels. The additional expert insert into the model when using the intermediate artificial ratings reduces the difficulty in learning the model parameters.

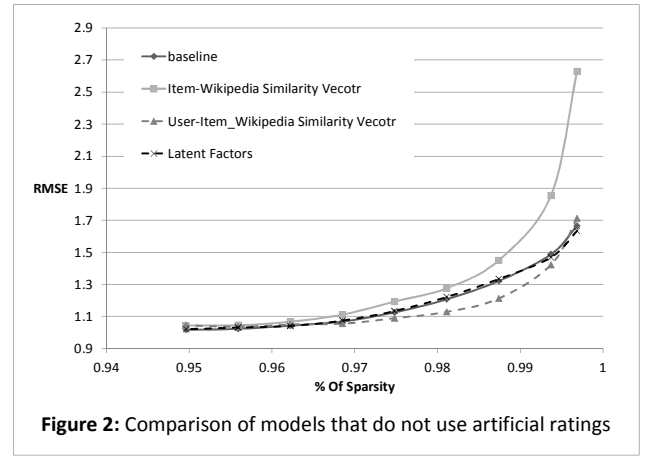

**Figure 2:** Comparison of models that do not use artificial ratings

**Table 1:** Improvement in % for models that use artificial ratings

| % of Sparsity | Diff. Step Size | Additional Parameters | Two Separate Models |
|---|---|---|---|
| 0.996858 | 35.8% | 39.35% | 33.74% |
| 0.993703 | 25.65% | 36.5% | 27.37% |
| 0.987406 | 19.17% | 29.18% | 19.8% |
| 0.981109 | 14.3% | 20.8% | 14.18% |
| 0.974812 | 10.8% | 16% | 10% |
| 0.968514 | 7.7% | 11.87% | 6.96% |
| 0.962217 | 6.7% | 10% | 5.3% |
| 0.95592 | 6.1% | 8.76% | 3.6% |
| 0.949623 | 6% | 7.78% | 3% |

**Table 2:** Improvement in % for models that do not use item-item Wikipeida similarity

| % of Sparsity | Item Similarity Vector | User-Item Similarity Vecotr | Latent Factors |
|---|---|---|---|
| 0.996858 | -36.65% | -2.74% | 1.90% |
| 0.993703 | -19.79% | 4.61% | 1.28% |
| 0.987406 | -8.90% | 9.00% | -0.91% |
| 0.981109 | -5.34% | 7.00% | -1.15% |
| 0.974812 | -5.70% | 3.30% | -0.74% |
| 0.968514 | -3.86% | 1.20% | -0.58% |
| 0.962217 | -2.33% | -0.69% | 0.16% |
| 0.95592 | -2.10% | -1.26% | -0.80% |
| 0.949623 | -2.64% | -2.60% | -0.38% |

We suggest two possible future research directions: the first is the representation of additional types of Wikipedia similarities in the model (in either separate parameters or by integrating them into existing ones); the second is the evaluation of the model on other domains such as books or music, in an attempt to determine the generics of the proposed method.

## 7. REFERENCES

1. Golub, G. and C. Reinsch, Singular value decomposition and least squares solutions. Numerische Mathematik, 1970. 14(5): p. 403-420.
2. Maltz, D. and K. Ehrlich, Pointing the way: active collaborative filtering, in Proceedings of the SIGCHI conference on Human factors in computing systems. 1995, ACM Press/Addison-Wesley Publishing Co.: Denver, Colorado, United States. p. 202-209.
3. Adomavicius, G. and A. Tuzhilin, Toward the next generation of recommender systems: a survey of the state-of-the-art and possible extensions. Knowledge and Data Engineering, IEEE Transactions on, 2005. 17(6): p. 734-749.
4. Katz, G., Ofek N., Shapira B., Rokach L., Shani G. , Using Wikipedia to boost collaborative filtering techniques, in Proceedings of the fifth ACM conference on Recommender systems. 2011, ACM: Chicago, Illinois, USA. p. 285-288.
5. Middleton, S.E., H. Alani, and D.D. Roure, Exploiting Synergy Between Ontologies and Recommender Systems. CoRR, 2002. cs.LG/0204012.
6. Semeraro, G., Lops P., Basile P., de Gemmis M.,, Knowledge infusion into content-based recommender systems, in Proceedings of the third ACM conference on Recommender systems. 2009, ACM: New York, New York, USA. p. 301-304.
7. Lee, J. W., S.G. Lee, and H.J. Kim, A probabilistic approach to semantic collaborative filtering using world knowledge. Journal of Information Science, 2011. 37(1): p. 49-66.
8. Bell, R.M. and Y. Koren, The Belkor 2008 soultion to the Netflix Prize. Available from http://www.netflixprize.com, 2008.
9. Ricci, F., L. Rokach, and B. Shapira, Introduction to Recommender Systems Handbook Recommender Systems Handbook, F. Ricci, et al., Editors. 2011, Springer US. p. 1-35.
10. Koren, Y., Factorization meets the neighborhood: a multifaceted collaborative filtering model, in Proceedings of the 14th ACM SIGKDD international conference on Knowledge discovery and data mining. 2008, ACM: Las Vegas, Nevada, USA. p. 426-434.
11. Kisilevich, S., Rokach, L., Elovici, Y., and Shapira, B.. Efficient multidimensional suppression for k-anonymity. IEEE Transactions on Knowledge and Data Engineering, 2010, 22(3), 334-347.
12. Jahrer, M., Tscher A. and Legenstein R.., Combining predictions for accurate recommender systems, in Proceedings of the 16th ACM SIGKDD international conference on Knowledge discovery and data mining. 2010, ACM: Washington, DC, USA. p. 693-702.
13. Salton, G. and C. Buckley, Term-weighting approaches in automatic text retrieva. Information Processing and Management, 1998. 24(5): p. 513-523.